# UniCross3D: Unified Cross-View and Cross-Domain Diffusion for Consistent Single-Image 3D Generation

U-Chae Jun†, Jaeeun Ko†, and Jiwoo Kang‡

Sookmyung Women's University, South Korea
{wjsdbco, rhwodms1223, jwkang}@sookmyung.ac.kr

**Abstract**

*Reconstructing detailed geometry and realistic appearance from a single RGB image is essential yet fundamentally challenging due to inherent ambiguities such as occlusion, lighting variations, and texture-geometry entanglement. While recent diffusion-based generative models have significantly improved novel view synthesis, existing approaches suffer from two critical limitations: lack of cross-view geometric consistency and insufficient cross-domain semantic alignment. To address these issues, we introduce* UniCross3D, *a unified cross-view and cross-domain diffusion framework designed explicitly for consistent and physically coherent 3D generation.* UniCross3D *features two novel contributions: (1) a cross-view latent regularization that enforces cross-view geometric consistency across synthesized viewpoints by penalizing latent variance, and (2) a cross-domain mutual information objective grounded in the physics of image formation, explicitly aligning synthesized color and normal maps. Extensive experiments demonstrate that* UniCross3D *achieves significantly improved view consistency and semantic alignment over state-of-the-art methods and yields higher-fidelity reconstructions, particularly under challenging textures and ambiguous viewpoints.*

## 1 Introduction

Single-image 3D generation, which reconstructs detailed geometry and realistic appearance from an RGB image, is a fundamental challenge in computer vision, with significant applications in augmented reality, virtual reality, and digital content creation [CXG*16; KLL21; PJBM23; LGT*23; HKK*24]. Although humans can effortlessly infer shape, texture, and spatial layout from a single photograph, replicating this ability computationally remains challenging due to inherent ambiguities, such as occlusion, lighting variation, and texture-geometry entanglement.

Recent advances in generative modeling, particularly diffusion models, have substantially improved 2D image synthesis and novel view generation [RBL*22; HJA20]. These capabilities have inspired a wave of single-image 3D generation approaches that leverage 2D generative priors to hallucinate plausible multiview images [LWV*23; HZG*24; LTZ*24; LGL*24]. A dominant paradigm among recent methods is the two-stage pipeline: first synthesizing multiview images or intermediate representations such as normals, then reconstructing a 3D shape using NeRF- or mesh-based optimization [QMH*24; LLL*24].

Despite their success, existing methods suffer from two major limitations. First, generated views often lack cross-view consistency: different synthesized viewpoints may depict objects with distorted geometry or mismatched appearance due to stochastic sampling or insufficient latent regularization [LLZ*24; LLL*24]. Fig. 1(a) demonstrates a typical failure scenario in existing methods, where an input image from a non-canonical viewpoint (*e.g.*, side-view or top-view) leads to significant geometric distortions and inconsistencies in the synthesized views. Second, when generating both color and normal images, most approaches treat these modalities as loosely coupled outputs, relying on shared latent codes or decoder attention but without physically grounded consistency constraints [LGL*24; WDL*23]. As a result, predicted color and geometry may not be semantically aligned: normals may not explain shading, and textures may not reflect surface orientation, especially in regions with complex materials or ambiguous structure. Fig. 1(b) illustrates this issue clearly, showing that existing approaches frequently struggle to disentangle intricate textures from underlying geometry, resulting in normals incorrectly reflecting texture-driven shading cues rather than actual surface geometry.

In this work, we propose UniCross3D, a unified cross-view and cross-domain diffusion framework for consistent 3D generation from a single image. Our method introduces two key innovations. First, we introduce a *multiview latent regularization* that pe-nalizes variance in the internal features of the denoising diffusion network across multiple noise samples and views, thus enforcing cross-view geometric consistency and semantic coherence in the latent space, as described in Fig. 1(a). Second, we propose a *cross-domain mutual information* maximization, grounded in the physics of image formation, which maximizes the statistical dependence between the synthesized color and normal maps. This explicitly enforces that geometry explains appearance, and vice versa, across all generated views, as shown in Fig. 1(b).

† These authors contributed equally to this work.
‡ Corresponding author.

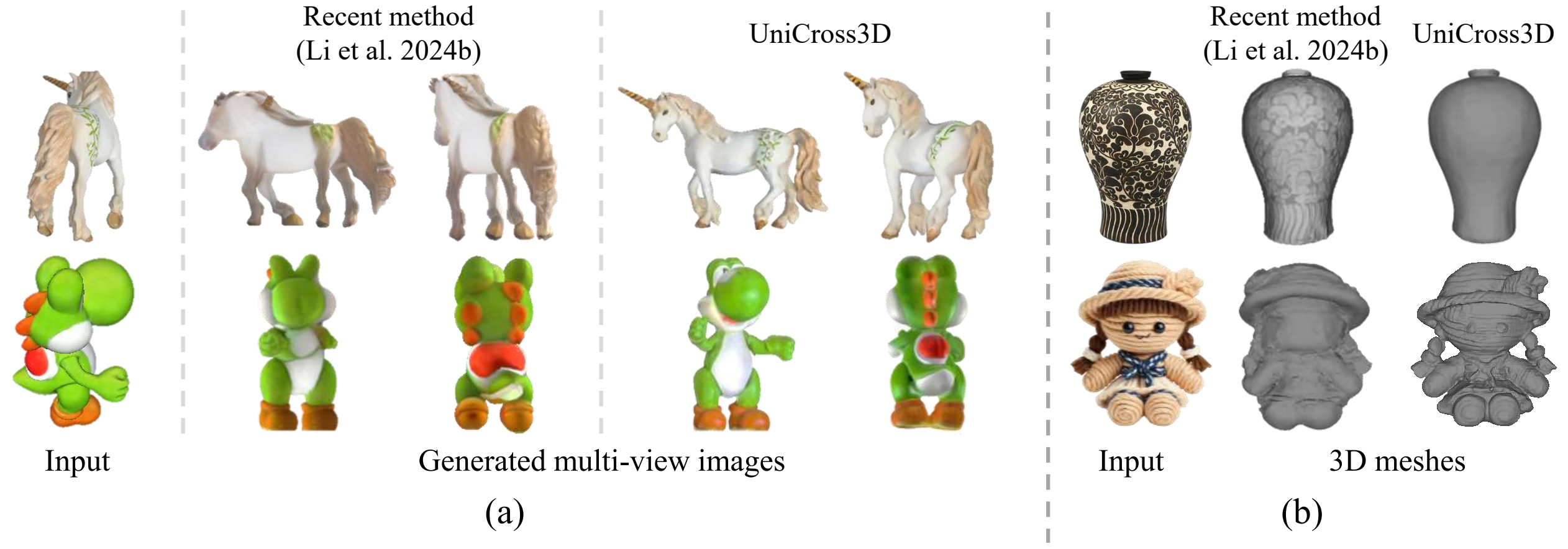

**Figure 1:** *Comparisons with the state-of-the-art 2D-to-3D method [LLL*24]. The proposed method,* UniCross3D, *addresses two fundamental limitations of the previous 2D-to-3D generation methods: (a) cross-view inconsistency and (b) cross-domain inconsistency. (a) Previous approaches suffer from geometric distortions and inconsistent appearances across synthesized views due to stochastic sampling and insufficient latent regularization. These issues become especially severe when generating views from novel, unseen viewpoints. (b) Existing methods lack strong consistency between color and normal maps, often misinterpreting texture patterns as geometry or producing shading that does not align with the actual surface normals. This results in semantic misalignments that degrade the physical plausibility of the reconstructed 3D shapes.*

These contributions are integrated into a conditional diffusion model that generates multiview RGB-normal pairs given a single image and target viewpoints. During training, we optimize the denoising loss jointly with cross-view and cross-domain consistency objectives, and at inference time, we decode geometry and texture from stable latent samples without optimization or post-processing.

Experimental results demonstrate that UniCross3D achieves superior view consistency and alignment of geometry and appearance compared to previous work. Our method produces more coherent geometry under complex shading and better preserves object identity and structure across views, leading to higher-fidelity 3D reconstructions.

## 2 Related Works

### 2.1 Single-Image 3D Generation

Single-image 3D generation aims to reconstruct a full 3D object from a single RGB image. Recent advances in diffusion models and neural rendering have enabled impressive results and can be broadly categorized into three paradigms: *optimization-based*, *feed-forward*, and *two-stage* methods. Each paradigm presents a trade-off between fidelity, inference speed, and supervision requirements.

*Optimization-based methods*, such as DreamFusion [PJBM23], Magic3D [LGT*23], and Hi3D [YCP*24], employ Score Distillation Sampling (SDS) to lift pretrained 2D diffusion priors into 3D representations through per-instance optimization. While these methods yield high-fidelity 3D results, they require minutes to hours per object, making them impractical for real-time applications. Moreover, many of these methods are designed for text-to-3D generation, which lacks direct conditioning on image inputs and often suffers from geometric ambiguity.

*Feed-forward methods* eliminate optimization per instance by training networks to directly regress a 3D representation from an image. LRM [HZG*24], Instant3D [LTZ*24], InstantMesh [XCG*24], and LGM [TCC*24] fall into this category, employing architectures such as triplanes [CLC*22] or Gaussian splatting [KKLD23] for real-time generation. These models can produce 3D assets in seconds but often suffer from resolution limitations due to compact latent encoding and constrained training scale. For example, InstantMesh adopts a mesh-based regression head with surface-level geometric supervision to improve quality, while LGM replaces triplanes with Gaussian splatting for better efficiency and higher-resolution synthesis.

*Two-stage generation pipelines* decompose the task into multiview synthesis followed by 3D reconstruction. Wonder3D [LGL*24], SyncDreamer [LLZ*24], Era3D [LLL*24], and Magic123 [QMH*24] represent this approach. These methods first generate multiview images and optionally normals using diffusion models and then apply mesh or volumetric reconstruction (*e.g.*, NeuS [WLL*21]). This paradigm balances flexibility and quality: multiview synthesis allows supervision on high-resolution views, while decoupled reconstruction supports accurate geometry modeling. However, challenges remain in achieving view-consistent generation, especially when normals and colors are modeled independently.

Moreover, two-stage approaches rely on mesh or volumetric based reconstruction backbones. The reconstruction process follows iterative optimization and progressively aligns generated multiview images into a single 3D representation. During this process, inconsistencies or noise across the input multiview images or normal maps propagate through the optimization procedure and hinder the accurate reconstruction of the desired 3D geometry and appearance.

Our method, UniCross3D, also adopts the two-stage approach but distinguishes itself by explicitly modeling cross-view and cross-domain consistency during multiview generation. By jointly learning the relationship between color and normal domains and enforcing latent coherence across views, we address a fundamental limitation of prior two-stage methods: the semantic and geometric inconsistency across generated views, enabling high-quality, consistent 3D mesh reconstruction from a single image with high fidelity.

### 2.2 Cross-View Consistency in 3D Generation

A key challenge in single-image 3D generation is ensuring consistency across multiple synthesized views. Since the input image provides only partial observations of the target object, the models must hallucinate plausible geometry and appearance for unseen viewpoints. Without appropriate constraints, this often leads to view-dependent inconsistencies, such as geometry drift, texture distortion, or semantic mismatch.

Early approaches such as Zero123 [LWV*23] introduced camera-conditioned 2D diffusion to synthesize novel views, but did not explicitly enforce consistency among them. More recent models address this by incorporating multiview attention or structural priors. For example, SyncDreamer [LLZ*24] introduces multiview-consistent generation by jointly denoising multiple views through cross-view self-attention, improving appearance alignment across viewpoints. Era3D [LLL*24] improves this with efficient row-wise attention for high-resolution multiview synthesis, achieving notable consistency on scale. Wonder3D [LGL*24] additionally leverages domain-switching diffusion to simultaneously predict color and normal maps from shared latent, contributing to more geometrically plausible results.

Other approaches also explore view alignment via architecture-level priors or dataset constraints. For example, LGM [TCC*24] employs a large multiview Gaussian prior to learn consistent geometry across generated views, and Magic123 [QMH*24] refines coarse 3D geometry using 2D and 3D diffusion stages that operate over multiple synthesized images. Unique3D [WLC*24] introduces geometric fusion techniques that aggregate multiview features into a unified 3D surface. Kiss3DGen [LYC*25] fine-tunes 2D Diffusion Transformer models (DiT) [PX23] to jointly predict color and normal maps from multiple viewpoints in a single forward generation, enhancing cross-view coherence through a unified multiview representation.

Despite these advances, consistency is often enforced implicitly via attention mechanisms, shared latent features, or reconstruction-level post-processing, without directly constraining the variance or stability of the underlying latent representations. This makes the output vulnerable to stochastic noise and sampling artifacts.

In contrast, UniCross3D introduces a *multiview latent regularization* mechanism that explicitly penalizes variance in bottleneck features of the denoising network between samples for the same viewpoint. This encourages the model to map different noise realizations into a consistent latent representation, thereby reducing shape hallucination, enhancing semantic coherence, and improving 3D reconstruction fidelity. Compared to attention-based consistency, our approach enforces alignment in the latent space itself, offering a complementary and more explicit mechanism for geometric stability.

To our knowledge, UniCross3D is the first work to explicitly propose a latent-level regularization objective that minimizes cross-view variation during multiview generation.

### 2.3 Cross-Domain Alignment of Appearance and Geometry

Generating coherent and physically consistent 3D objects from a single image requires aligning appearance (*e.g.*, RGB colors) and geometry (*e.g.*, surface normals). Without explicit constraints, synthesized geometry and appearance may appear individually plausible yet remain semantically and physically misaligned, especially in complex textural or shading scenarios.

Recent methods such as Wonder3D [LGL*24] and Era3D [LLL*24] explicitly synthesize both color and normal maps, typically employing cross-domain attention mechanisms or shared decoders to implicitly couple these modalities. However, these approaches rely predominantly on indirect or data-driven supervision, which lacks explicit physical constraints on geometry-appearance relationships. Similarly, Score Jacobian Chaining [WDL*23] refines geometry predictions guided by pre-trained 2D diffusion models but does not explicitly enforce a consistency constraint between predicted shading and normals.

On the other hand, triplane-based methods such as LRM [HZG*24] and Instant3D [LTZ*24] directly regress both geometry and texture from a shared latent representation trained under volumetric rendering supervision. These methods naturally inherit latent-level coherence between geometry and appearance. However, this consistency remains implicitly driven by data supervision and does not necessarily enforce physically meaningful alignment between surface normals and observed shading. In addition, their compact triplane-based architecture often restricts output resolution, which can further hinder accurate shading cues or fine-grained surface detail. As a result, such implicit approaches might struggle with generalizing beyond seen scenarios, particularly when faced with complex materials, ambiguous shading patterns, or texture-driven illusions that were not captured in the training distribution.

In contrast, UniCross3D explicitly addresses these limitations by incorporating a physics-inspired, information-theoretic constraint through cross-domain mutual information maximization. Under the assumption of Lambertian reflectance, the observed color is fully determined by the surface normals and lighting conditions [ZTCS99; BM15]. We leverage this principle explicitly through a conditional mutual information objective, ensuring that the generated appearance and geometry not only share latent features but also explain each other physically. Consequently, UNICROSS3D achieves robust and semantically consistent geometry-appearance alignment, especially under challenging textures and complex shading scenarios that previous implicit approaches often did not handle effectively.

## 2.4 Physics-based Approaches in 3D Generation

Recent 3D generation approaches explicitly incorporate physics-based constraints into the generation pipeline to ensure the physical plausibility of reconstructed 3D assets.

SF3D [BHVJ25] and ARM [FYB*25] adopt a disentangled modeling approach that decouples appearance from geometry during the multiview generation process for 3D reconstruction. Rather than directly predicting view-dependent colors conditioned on geometry, these methods predict appearance by decomposing it into independent physical properties, including albedo, roughness, and metallic properties.

These disentangled attributes are integrated through a differentiable rendering process [SGY*21].

The generation process is then enforced by comparing this recomposed output with the original input image to maintain photometric consistency. This structural constraint prevents the network from generating physically implausible attributes and enables 3D assets to maintain stable appearance under varying lighting conditions.

However, these approaches primarily focus on appearance-level constraints and lack explicit regularization for enforcing cross-view and cross-domain consistencies.

In contrast, UNICROSS3D leverages a Lambertian reflectance-based constraint to reinforce cross-view coherence between color and surface normal during generation. Rather than constraining appearance attributes alone, our method explicitly regularizes the physical relationship between color and geometry across views through a cross-domain mutual information objective.

This enables stable alignment between appearance and surface geometry across viewpoints, ensuring cross-view and cross-domain consistencies within a unified generative framework.

# 3 Methodology

## 3.1 Pipeline Overview

Figure 2 provides an overview of the proposed pipeline. UNICROSS3D takes a single-view image as input and generates multiview data from diverse viewpoints, which is then used to reconstruct a 3D textured mesh.

Given a single input image, UNICROSS3D employs a unified cross-view and cross-domain diffusion framework to synthesize color images and corresponding normal maps from $N$ novel viewpoints ($N = 6$ in our implementation). During training, the model minimizes the joint objective in Eq. (18), which incorporates both cross-domain and cross-view consistency terms into the denoising diffusion objective, $\mathcal{L}_{\text{diffusion}}$.

The first term, $\mathcal{L}_{\text{Var}}$, minimizes the latent variance across multiview samples. By enforcing stability in the internal representations of different viewpoints, this regularization prevents the model from overfitting to specific input views and ensures consistent generation even when the input viewpoint varies. This mechanism is crucial for maintaining geometric coherence and structural fidelity, particularly under challenging or unseen viewpoints. More details on this multiview latent regularization are discussed in Sec. 3.3.

The second term, $\mathcal{L}_{\text{MI}}$, maximizes mutual information between the color and normal maps generated at each viewpoint. This encourages semantic alignment across domains, allowing the model to learn complementary representations of appearance and geometry. As a result, UNICROSS3D is able to disentangle shading from the structure and preserve fine surface details without introducing artifacts or inconsistencies across modalities. A detailed explanation of this cross-domain consistency mechanism is provided in Sec. 3.4.

Through the combination of these two consistency objectives, UNICROSS3D learns a semantically grounded relationship between appearance and geometry. It is particularly effective in complex scenarios where texture and geometry are entangled, such as objects with intricate patterns, fine-grained structure, or material variation. Furthermore, by reducing dependence on specific input viewpoints, the model generalizes well to unseen angles or objects with limited observable cues, enabling robust and consistent multiview generation.

Finally, the generated multiview color images and normal maps are used to reconstruct a 3D textured mesh using a NeRF-based method, in our case, NeuS [WLL*21]. This reconstruction takes advantage of the complementary strengths of color and normal cues for surface optimization. Details on this reconstruction process can be found in Sec. 3.6.

## 3.2 Problem Formulation

We denote a single RGB image captured under perspective projection as $y \in \mathbb{R}^{H\times W\times 3}$. The objective is to generate a set of $N$ views $\{\pi_i\}_{i=1}^{N}$, each producing a pair of a color image $x^{(i)} \in \mathbb{R}^{H\times W\times 3}$ and a normal map $n^{(i)} \in \mathbb{R}^{H\times W\times 3}$. Although the ground-truth color and normal maps $\{x^{(i)}, n^{(i)}\}$ are available during training, the diffusion model is trained not to predict them directly but to denoise a latent variable $z_0^{(i)}$ that implicitly encodes them. Therefore, given $y$ and target viewpoints $\{\pi_i\}_{i=1}^{N}$, the joint conditional distribution over outputs is defined as

$$p_\theta\big(\{x^{(i)}, n^{(i)}\}_{i=1}^{N} \;\big|\; y, \{\pi_i\}_{i=1}^{N}\big). \tag{1}$$

For a generic viewpoint $\pi$, we factorize the conditional joint over colors and normals as

$$\begin{aligned} p_\theta(x, n \mid y, \pi) &= p_\theta\big(n \mid y, \pi\big)\; p_\theta\big(x \mid n, y, \pi\big) &(2)\\ &= p_\theta\big(x \mid y, \pi\big)\; p_\theta\big(n \mid x, y, \pi\big). &(3)\end{aligned}$$

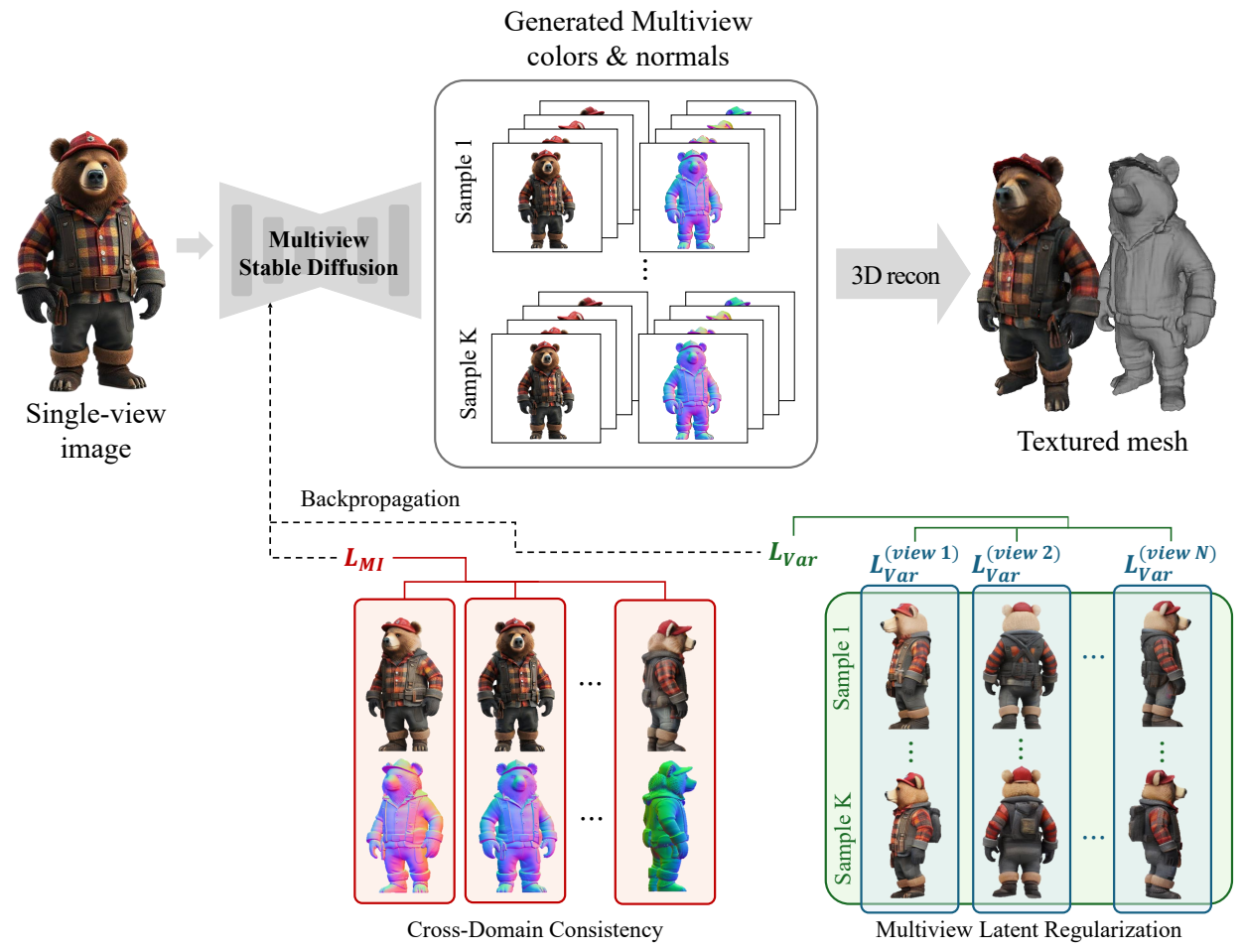

**Figure 2:** *Overview of* UNICROSS3D *framework.*

These factorizations follow from Bayes' rule and reflect two valid decompositions of the same joint conditional distribution, highlighting different modeling choices for supervision or auxiliary constraints. We will leverage this factorization in Sec. 3.4 to define an information-theoretic consistency loss between the two domains. Since we generate all $N$-view pairs jointly, Eq. (1) becomes

$$p_\theta\big(\{x^{(i)}, n^{(i)}\}_{i=1}^N \mid y\big) = \prod_{i=1}^{N} p_\theta\big(x^{(i)}, n^{(i)} \mid y, \pi_i\big). \tag{4}$$

To capture each factor $p_\theta(x^{(i)}, n^{(i)} \mid y, \pi_i)$, we introduce a latent variable $z_0^{(i)} \in \mathbb{R}^d$ that encodes the pair $(x^{(i)}, n^{(i)})$. A forward diffusion process perturbs $z_0^{(i)}$ through $T$ steps as

$$q\big(z_t^{(i)} \mid z_{t-1}^{(i)}\big) = \mathcal{N}\big(\sqrt{1-\beta_t}\, z_{t-1}^{(i)}, \beta_t I\big), \quad t = 1, \dots, T, \tag{5}$$

$$z_t^{(i)} = \sqrt{\bar\alpha_t}\, z_0^{(i)} + \sqrt{1-\bar\alpha_t}\,\epsilon, \quad \epsilon \sim \mathcal{N}(0, I), \bar\alpha_t = \prod_{s=1}^{t} \alpha_s, \tag{6}$$

where $\alpha_t = 1 - \beta_t$. After $T$ steps, $z_T^{(i)}$ is approximately standard Gaussian. The reverse (denoising) distribution is parameterized by $\epsilon_\theta$:

$$p_\theta\big(z_{t-1}^{(i)} \mid z_t^{(i)}, y, \pi_i\big) = \mathcal{N}\big(\mu_\theta(z_t^{(i)}, y, \pi_i, t), \sigma_t^2 I\big), \tag{7}$$

$$\mu_\theta\big(z_t^{(i)}, y, \pi_i, t\big) = \frac{1}{\sqrt{\alpha_t}}\Big(z_t^{(i)} - \frac{\beta_t}{\sqrt{1-\bar\alpha_t}}\,\epsilon_\theta\big(z_t^{(i)}, y, \pi_i, t\big)\Big). \tag{8}$$

Training minimizes the expected noise prediction error across all $N$ views:

$$\mathcal{L}_{\text{diffusion}} = \sum_{i=1}^{N} \mathbb{E}_{z_0^{(i)}, t, \epsilon} \Big\| \epsilon - \epsilon_\theta\big(z_t^{(i)}, y, \pi_i, t\big) \Big\|_2^2, \tag{9}$$

where $z_t^{(i)}$ follows Eq. (6), $\epsilon \sim \mathcal{N}(0, I)$, and $z_0^{(i)}$ is the encoding of the ground-truth pair $(x^{(i)}, n^{(i)})$. A decoder then maps each recovered $z_0^{(i)}$ to $(x^{(i)}, n^{(i)})$. During inference, for each $\pi_i$, we sample $z_T^{(i)} \sim \mathcal{N}(0, I)$ and iteratively apply the denoiser in Eq. (7) conditioned on $(y, \pi_i)$ to obtain $z_0^{(i)}$. The decoder finally produces the color-normal pair $(x^{(i)}, n^{(i)})$.

The conditional diffusion model offers a framework for generating multiview color and normal images from a single input; however, it cannot ensure semantic coherence across stochastic samples, geometric accuracy under perspective distortion, and cross-domain consistency. To address these limitations, we introduce key contributions: a latent-space variance regularization (Sec. 3.3) to maintain coherence across multiple samples from the same input and a cross-domain mutual information constraint (Sec. 3.4) for aligning color and normal predictions, collectively ensuring geometric and semantic consistency across generated views.

### 3.3 Multiview Latent Regularization

In single-view conditioned 3D generation, predicting all novel views from a single image is severely under-constrained. Even when conditioned on the same input $y$ and viewpoint $\pi$, a diffusion model can produce various latent samples $\{z_{0,k}^{(i)}\}_{k=1}^K$ in a given view $\pi_i$. Although some variability is acceptable, since slight texture or shading changes across stochastic draws can be natural, excessive variance in the bottleneck features can lead to geometric or semantic drift. In particular, if the internal U-Net representation for two samples differs drastically, their decoded color-normal pairs $(x_k^{(i)}, n_k^{(i)})$ can imply inconsistent object shapes or lighting, reducing multiview coherence.

To mitigate this inconsistency, we assume that latent representations obtained from $N$ different viewpoints should preserve a consistent semantic structure across viewpoints and lie in a shared latent manifold.

To enforce this assumption, we penalize feature-space variance at a designated "bottleneck" layer of the U-Net. Denote by $f_\theta^{(i)}(y, \pi_i, t) \in \mathbb{R}^D$ the vectorized feature output of the U-Net's central (lowest-resolution) layer when denoising $z_t^{(i)}$ at timestep $t$. For $K$ independent noise realizations $\epsilon_k \sim \mathcal{N}(0, I)$, let

$$z_{t,k}^{(i)} = \sqrt{\bar\alpha_t}\, z_0^{(i)} + \sqrt{1-\bar\alpha_t}\,\epsilon_k, \quad k = 1, \dots, K, \tag{10}$$

and let $f_k^{(i)} = f_\theta^{(i)}(y, \pi_i, t; z_{t,k}^{(i)})$ be the corresponding bottleneck feature for sample $k$. This variance reflects the model's uncertainty over the latent representation $f^{(i)}$ when denoising different noise realizations $z_{t,k}^{(i)}$. From a Bayesian perspective, this corresponds to the posterior variance under the model's learned distribution. Minimizing this variance encourages posterior concentration [GG16; FHL19], promoting semantic stability across stochastic samples. We therefore compute the sampled posterior variance:

$$\mathrm{Var}\big(f^{(i)}\big) = \frac{1}{K}\sum_{k=1}^{K} \Big\| f_k^{(i)} - \mu_f^{(i)} \Big\|_2^2, \; \mu_f^{(i)} = \frac{1}{K}\sum_{k=1}^{K} f_k^{(i)}. \tag{11}$$

We then introduce a variance-penalty term for view $\pi_i$:

$$\mathcal{L}_{\text{Var}}^{(i)} = \mathrm{Var}\big(f^{(i)}\big). \tag{12}$$

Summing over all $N$ views yields the total regularization:

$$\mathcal{L}_{\text{Var}} = \sum_{i=1}^{N} \mathcal{L}_{\text{Var}}^{(i)}. \tag{13}$$

By minimizing $\mathcal{L}_{\text{Var}}$, we encourage the U-Net's latent features to remain centered and compact under different noise perturbations. This prevents semantic drift: although different draws $k$ can still produce slight view-dependent variation, their core representation remains stable, ensuring that the decoded $(x_k^{(i)}, n_k^{(i)})$ remain geometrically and visually consistent for each $\pi_i$. In practice, $\mathcal{L}_{\text{Var}}$ is generated at a single randomly chosen timestep $t$ per iteration (*e.g.*, uniform $t \in \{1, \ldots, T\}$) to reduce computation, and does not require ground-truth pairs. Combined with the diffusion loss in Eq. (9) and the mutual-information loss in Eq. (17), this latent-space regularizer completes the set of consistency constraints applied during training.

### 3.4 Cross-Domain Consistency

We seek to enforce the consistency between $x$ and $n$ by grounding our model in the physics of image formation.

Specifically, assuming a Lambertian shading model, the observed color is represented as a product of surface albedo $A$ and shading $S(n, \ell)$:

$$x = A \cdot S(n, \ell). \tag{14}$$

The surface normal $n$ and lighting $\ell$ determine shading and thereby color. Thus, predicting $x$ from $n$ is a forward rendering process, while predicting $n$ from $x$ constitutes a decomposition, *i.e.*, an ill-posed inverse problem that can be ambiguous without additional priors [ZTCS99; BM15]. This asymmetry motivates us to treat $n$ as a latent explanatory factor for $x$, and to ensure that both domains align semantically during generation. In particular, we observe that the conditional joint distribution $p_\theta(x, n \mid y, \pi)$ permits two distinct factorizations, as shown in Eqs. (2) and (3), reflecting different modeling assumptions. If $x$ and $n$ were conditionally independent given $y$ and $\pi$, both factorizations would degenerate into marginals. The fact that both factorizations are valid yet different implies a nontrivial statistical dependency between $x$ and $n$, which motivates us to introduce mutual information as an explicit regularization term [BBR*18].

Given the model-defined conditional distributions $p_\theta(x, n \mid y, \pi)$ and marginals $p_\theta(x \mid y, \pi)$, $p_\theta(n \mid y, \pi)$, we define the conditional mutual information as

$$\mathcal{I}(x; n \mid y, \pi) = \mathbb{E}_{(x,n) \sim p_\theta(x,n|y,\pi)} \left[ \log \frac{p_\theta(x, n|y, \pi)}{p_\theta(x|y, \pi)\, p_\theta(n|y, \pi)} \right]. \tag{15}$$

Maximizing $\mathcal{I}(x; n)$ encourages to encode complementary information in $x$ and $n$, ensuring that the color image can explain the corresponding normal map and vice versa. This is especially important in our setup, where both outputs are generated from shared latent variables, and the model must resolve their geometric correspondence from a single input.

Because Eq. (15) is intractable to compute directly, we adopt the InfoNCE estimator [vdOLV18], which is a lower bound on conditional mutual information [POvdO*19]. For each novel view $\pi_i$, let $\{(x_k^{(i)}, n_k^{(i)})\}_{k=1}^{K}$ be $K$ independent samples from the model conditioned on the same $y$. We define feature embeddings $f_\phi^x(x)$ and $f_\phi^n(n)$ via small projection heads. The mutual information loss between cross domains for view $\pi_i$ is

$$\mathcal{L}_{\text{MI}}^{(i)} = -\frac{1}{K} \sum_{k=1}^{K} \log \frac{\exp\left(\text{sim}\left(f_\phi^x(x_k^{(i)}), f_\phi^n(n_k^{(i)})\right)/\tau\right)}{\sum_{j=1}^{K} \exp\left(\text{sim}\left(f_\phi^x(x_k^{(i)}), f_\phi^n(n_j^{(i)})\right)/\tau\right)}, \tag{16}$$

where $\text{sim}(u, v) = u^\top v / (\|u\| \|v\|)$ is the cosine similarity and $\tau$ a temperature. To apply the contrastive loss, we treat each color-normal pair $(x_k^{(i)}, n_k^{(i)})$ as a positive pair, and define the negatives as mismatched color-normal pairs $(x_k^{(i)}, n_j^{(i)})$ with $j \neq k$ but drawn from the same diffusion timestep $t$ and using the same noise realization $\epsilon$. This ensures that all samples are generated under the same stochastic condition, and any discrepancy in their embeddings arises solely from the underlying content mismatch between different object instances [HFL*19].

Summing over all $N$ views yields

$$\mathcal{L}_{\text{MI}} = \sum_{i=1}^{N} \mathcal{L}_{\text{MI}}^{(i)}. \tag{17}$$

This loss forces each pair $(x_k^{(i)}, n_k^{(i)})$ to remain closer in the embedding space than any mismatched pair $(x_k^{(i)}, n_j^{(i)})$ for $j \neq k$, thereby aligning the generated colors and normals at every viewpoint. Integrating $\mathcal{L}_{\text{MI}}$ with the diffusion objective of Eq. (9) ensures a physically grounded and consistent generation. In practice, we back-propagate through the denoiser $\epsilon_\theta$ to compute $\{x_k^{(i)}, n_k^{(i)}\}$ at each diffusion step $t$ but only apply $\mathcal{L}_{\text{MI}}$ after decoding $z_0^{(i)}$.

### 3.5 Training Objective

We train the denoising model $\epsilon_\theta$ by minimizing a combined loss $\mathcal{L}_{\text{denoising}}$ that integrates the denoising objective in Eq. (9) and the latent-space regularization term in Eq. (13):

$$\mathcal{L}_{\text{denoising}} = \mathcal{L}_{\text{diffusion}} + \lambda_{\text{Var}} \mathcal{L}_{\text{Var}} + \lambda_{\text{MI}} \mathcal{L}_{\text{MI}}, \tag{18}$$

where $\lambda_{\text{Var}}$ and $\lambda_{\text{MI}}$ are weighting hyperparameters. Minimizing this objective enables the model to reconstruct geometrically consistent and semantically aligned color-normal pairs while explicitly correcting for instance-specific perspective distortion. At inference, for each viewpoint $\pi_i$, we sample $z_T^{(i)} \sim \mathcal{N}(0, I)$ and iteratively apply the reverse diffusion steps in Eq. (7) and (8), conditioned on the input image $y$ and viewpoint $\pi_i$. The final decoded latent $z_0^{(i)}$ yields the output pair $(x^{(i)}, n^{(i)})$ for view $\pi_i$.

### 3.6 3D Reconstruction

We reconstruct a 3D mesh from the generated multiview color images and normal maps using a Signed Distance Field (SDF)-based method, specifically NeuS [WLL*21]. During training, the multiview color images supervise the radiance field by minimizing the mean squared error (MSE) between the predicted color $\hat{C}(r)$ and the ground-truth rendered color $C(r)$ for each ray $r$. The color loss

$L_{\text{rgb}}$ is defined over the set of sampled pixels $\mathcal{P}$ as follows:

$$L_{\text{rgb}} = \frac{1}{|\mathcal{P}|} \sum_{p \in \mathcal{P}} \left\| \hat{C}(r_p) - C(r_p) \right\|_2^2. \quad (19)$$

Unlike conventional SDF-based methods that rely solely on color supervision, we further leverage the predicted multiview normal maps as geometric supervision. Specifically, we encourage alignment between the predicted surface normals $n_p$ and the SDF gradient $\hat{n}_p$, thus promoting more accurate surface orientation and improving geometric fidelity. This is achieved by minimizing the cosine distance between them via the normal loss:

$$L_{\text{normal}} = \frac{1}{|\mathcal{P}|} \sum_{p \in \mathcal{P}} \Big(1 - \cos\big(\hat{n}_p, n_p\big)\Big), \quad (20)$$

where $\cos(\cdot,\cdot)$ denotes the cosine similarity function.

The final training objective $L_{\text{recon}}$ for reconstruction combines both losses:

$$L_{\text{recon}} = L_{\text{rgb}} + L_{\text{normal}}. \quad (21)$$

By incorporating both color and normal supervision from multiview inputs, the proposed method captures fine-grained geometry and texture details more effectively. This joint supervision enables the generation of sharper, more accurate, and visually coherent 3D meshes, particularly in challenging regions with complex surface variation or high-frequency appearance features.

## 4 Experiments

In this section, we conducted experiments to evaluate the effectiveness of the proposed method in various tasks and settings. Specifically, Sec. 4.2 presents comprehensive comparisons with state-of-the-art methods on 2D-to-3D generation [LXJ*23; LGL*24; HZG*24; WWC*24], text-to-3D generation [WWC*24; JZH*24; MWZ*24], and single view-to-multiview generation [LLL*24; LGL*24; LLZ*24]. Sec. 4.3 reports ablation studies that isolate the contributions of multiview latent regularization and cross-domain consistency objectives.

Furthermore, Sec. 4.4 analyzes the capabilities of the proposed framework by evaluating the surface fidelity of reconstructed geometry and the robustness of multiview generation to variations in input viewpoints. Finally, Sec. 4.5 presents a sensitivity analysis of the proposed method with respect to key hyperparameters, including $\lambda_{\text{MI}}$ and $\lambda_{\text{Var}}$ in (18).

### 4.1 Experimental Details

#### 4.1.1 Training Datasets.

For training the proposed method, we used a subset of Objaverse [DSS*23] and rendered 96 images per 3D object at a resolution of 512×512. Specifically, we first divided the azimuth range from 0° to 360° into 16 evenly spaced viewpoints, fixing the elevation at 0°, and rendered 16 images using an orthographic camera. Then, for each azimuth angle, we randomly sampled an elevation angle from the range $[-20°, 40°]$ and rendered 5 images using a perspective camera. For perspective rendering, the focal length was randomly selected from {35, 50, 85, 105, 135} mm. This hybrid sampling strategy ensures both view diversity and consistent alignment, enabling the model to learn robust geometric priors from both canonical and challenging viewpoints.

To correct scale mismatches arising from structural differences between orthographic and perspective projection, as well as scale changes induced by focal length variation in perspective cameras, we apply distance-based scale compensation and focal-length normalization. In particular, perspective projection causes scale variations in the rendered image depending on the focal length $f$, even for the same object, potentially introducing bias during training. To address this, we compute the camera distance $d$ that maintains a consistent object scale across views. Given a focal length $f$ and a fixed orthographic scale $s$, we set the camera distance to $d = f/s$. This distance is used as the offset between the object center and the camera center during rendering, effectively aligning the apparent size of objects across different projection types. In our experiments, we randomly sample perspective views using a discrete set of focal lengths: $f = \{35, 50, 85, 105, 135\}$ mm. To mitigate instability due to the wide range of focal values, we normalize the focal length by defining a dimensionless quantity $\tilde{f} = f/35$, where 35 mm is the smallest focal length in our set. This normalized focal length $\tilde{f}$ is used as an input conditioning signal during training. For orthographic views that do not employ focal length, we simply set $\tilde{f} = 0$. By applying distance-based scale compensation and normalized focal length conditioning, we reduce data inconsistency introduced by diverse camera intrinsics and projection types. This normalization strategy allows the model to generalize better across varying viewpoints and camera configurations, leading to more stable and accurate learning.

#### 4.1.2 Performance Metrics.

We evaluated UniCross3D on two key aspects of 2D-to-3D generation: *novel view synthesis* and *3D reconstruction*. For novel view synthesis, we use Peak Signal-to-Noise Ratio (PSNR), Structural Similarity Index Measure (SSIM) [WBSS04], and Learned Perceptual Image Patch Similarity (LPIPS) [ZIE*18]. For 3D reconstruction, we follow prior works [LWV*23; LGL*24] and use Chamfer Distance (CD) and volumetric Intersection over Union (IoU). Additionally, for the text-to-3D generation task, we evaluated the realism and semantic alignment of the generated results using Fréchet Inception Distance (FID) [HRU*17] and CLIP score [RKH*21].

#### 4.1.3 Implementation Details.

UniCross3D generates a 3D mesh from a single-view input image using a two-stage pipeline: (1) it first synthesizes multiview color images and normal maps from the given view, and (2) it then performs 3D reconstruction from these color-normal pairs using Eq. (21). In our implementation, we generate six multiview pairs corresponding to a fixed elevation of 0° and azimuth angles $\beta+30°$, $\beta+90°$, $\beta+150°$, $\beta+210°$, $\beta+270°$, and $\beta+330°$, where $\beta$ is the azimuth of the input image. We set $\lambda_{\text{MI}} = 0.05$ and $\lambda_{\text{Var}} = 0.01$ in Eq. (18), which provided the best performance across all experiments. For a more detailed sensitivity analysis, please refer to Sec. 4.5. Our multiview diffusion model is trained on top of the open-source text-to-image model SDXL [PEL*23]. Following the Era3D [LLL*24] protocol, we train the SDXL model to perform multiview generation conditioned on camera viewpoints, jointly synthesizing color maps and normal maps from a single input image. The model is trained for 60K steps over four days using 8 NVIDIA A100 GPUs, with a batch size of 192. The learning rate is initialized at 1e−4 and decayed to 1e−5 after 10K steps. During inference, we use 50 denoising steps with a DDIM sam-pler [SME21] and apply classifier-free guidance [HS22] with a scale of 3.0.

**Table 1:** *Performance comparisons with multiview images and textured mesh generated by 2D-to-3D methods.*

| Method | PSNR (↑) | SSIM (↑) | LPIPS (↓) | CD (↓) | IoU (↑) |
|---|---|---|---|---|---|
| One-2-3-45 | 16.1058 | 0.8874 | 0.1812 | 0.0313 | 0.4142 |
| SyncDreamer | 18.2132 | 0.8462 | 0.1572 | 0.0283 | 0.4482 |
| Wonder3D | 18.0932 | 0.8995 | 0.1536 | 0.0261 | 0.4663 |
| Era3D | 19.1325 | 0.9124 | 0.1486 | 0.0153 | 0.5217 |
| CRM | 18.4407 | 0.9088 | 0.1366 | 0.0141 | 0.5218 |
| OpenLRM | 18.0433 | 0.8957 | 0.1560 | 0.0336 | 0.3947 |
| Kiss3DGen | 18.9853 | 0.9103 | 0.1397 | 0.0148 | 0.5169 |
| UNICROSS3D (Ours) | **21.8354** | **0.9642** | **0.1061** | **0.0125** | **0.6125** |

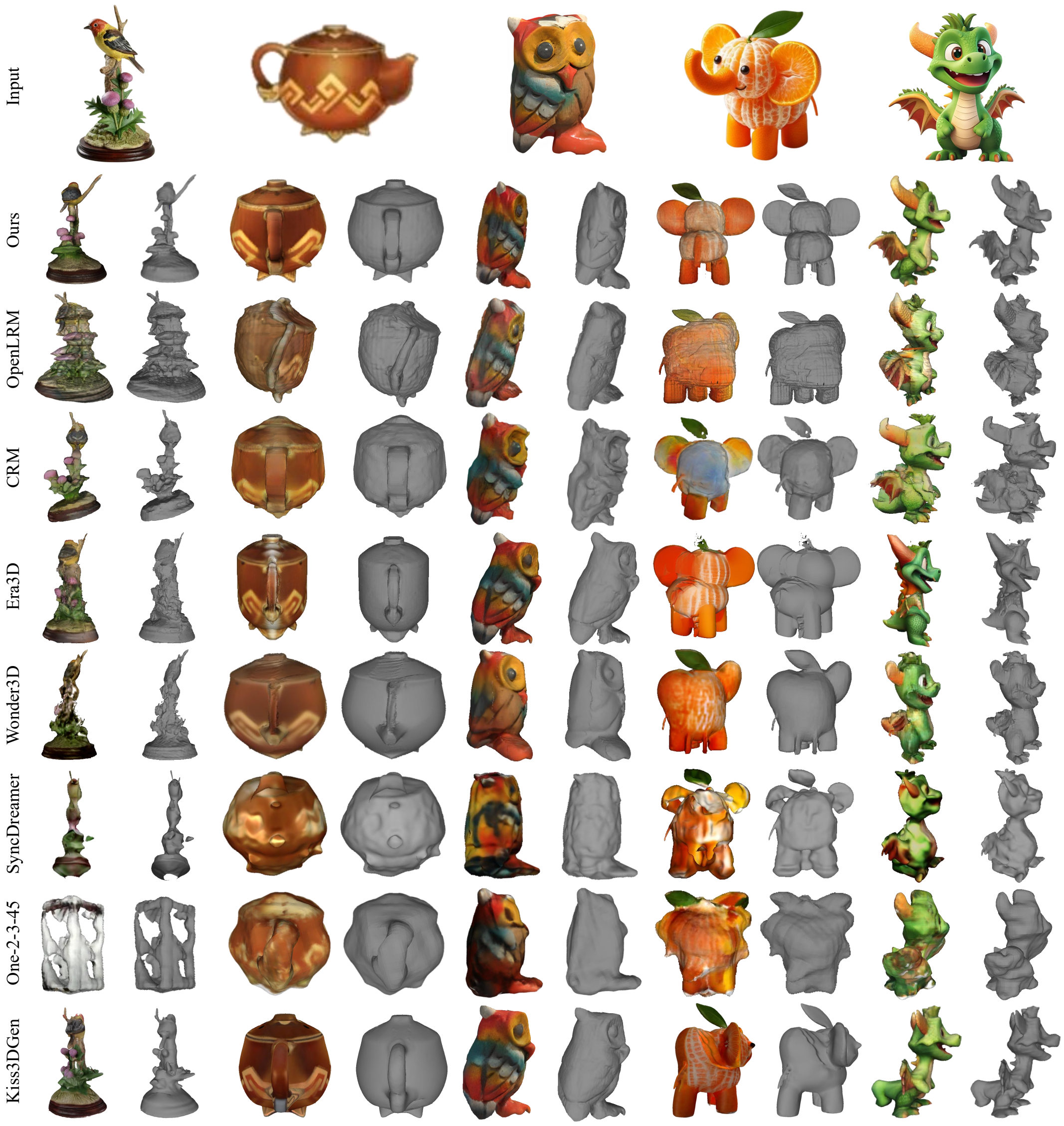

**Figure 3:** *Qualitative comparisons of 3D reconstruction results with state-of-the-art 2D-to-3D methods.*

### 4.2 Comparisons with State-of-the-Art

In this section, we evaluated the performance of the proposed UNICROSS3D framework through comprehensive comparisons with state-of-the-art baselines. Specifically, in Sec. 4.2.1, we evaluated the performance of UNICROSS3D on the 2D-to-3D generation task by comparing it with recent image-to-3D methods [LXJ*23; LGL*24; HZG*24; WWC*24; LYC*25]. These baselines represent a variety of paradigms, including feed-forward regression, triplane-based models, and cross-domain diffusion frameworks. In Sec. 4.2.2, we evaluated the generality of UNICROSS3D by integrating it with off-the-shelf text-to-image diffusion models [PEL*23], enabling a plug-and-play text-to-3D generation pipeline. We compared our performance with recent text-to-3D approaches [WWC*24; JZH*24; MWZ*24]. Additionally, in Sec. 4.2.3, we compared UNICROSS3D with state-of-the-art single image-to-multiview generation methods [LLL*24; LGL*24; LLZ*24] to evaluate the effectiveness of the proposed framework.

#### 4.2.1 2D-to-3D Generation

To validate the performance of UNICROSS3D, we compared the performance of recent 2D-to-3D methods [LXJ*23; LGL*24; LLZ*24; HZG*24; WWC*24; LLL*24; LYC*25]: One-2-3-45 [LXJ*23], Wonder3D [LGL*24], Era3D [LLL*24], SyncDreamer [LLZ*24], OpenLRM [HZG*24], CRM [WWC*24], and Kiss3DGen [LYC*25]. These methods represent a range of paradigms, from diffusion-based view synthesis to triplane-based direct reconstruction. One-2-3-45 generates multiview images using Zero123, then reconstructs 3D surfaces with an SDF-based network like SparseNeuS. Wonder3D creates multiview normal maps and color images through cross-domain diffusion, followed by geometry-aware normal fusion. Era3D produces multiview images by predicting camera parameters from a single input and using epipolar priors via row-wise attention. SyncDreamer employs a synchronized multiview diffusion framework, modeling the joint distribution of views with a shared noise predictor and 3D-aware conditioning to generate view-consistent images from a single input. OpenLRM encodes the image using self-distillation (DINO) [CTM*21], projects features onto triplane representations with a transformer decoder, and renders outputs via NeRF-based MLPs. CRM builds triplane features from diffusion-generated orthographic views and canonical coordinate maps using a convolutional U-Net, reconstructing textured meshes through MLP decoding and Flexicubes.

Kiss3DGen fine-tunes a pretrained 2D image diffusion model to generate a tiled 3D bundle image consisting of multiview color images and normal maps. For evaluation, we used 500 distinct single-view input images randomly sampled from the Google Scanned Objects (GSO) dataset [DFK*22], featuring diverse real-world objects captured under varying viewpoints and lighting conditions.

Quantitative comparisons are presented in Table 1. The results indicate that UNICROSS3D achieves the best empirical performance across all evaluation metrics. In particular, the improvements in CD and IoU reflect significant gains in 3D reconstruction quality. These improvements suggest that our unified cross-view and cross-domain diffusion framework enables the model to effectively capture geometric details and texture, while generating high-frequency multiview images and normal maps that are structurally consistent across viewpoints. As a result, the reconstructed meshes exhibit high fidelity and fine-grained accuracy.

Figure 3 provides visual comparisons of the reconstructed meshes, highlighting the robustness of UNICROSS3D under challenging viewpoints and object geometries. In the first column of Fig. 3, UNICROSS3D preserves the global geometry of the object while maintaining fine structural details such as birds overlapping with branches and flowers with sharp contours, even from rear views. In contrast, methods such as OpenLRM, CRM, SyncDreamer, and One-2-3-45 struggle to maintain the object's structure: the overall shape becomes distorted, and boundaries between flowers, branches, and birds appear blurred, indicating a lack of structural fidelity. Similarly, Wonder3D and Era3D tend to produce results where branches unnaturally intersect the bird's body or flower structures become oversmoothed and noisy. This trend continues with objects exhibiting complex texture, as seen in the fourth column of Fig. 3. UNICROSS3D preserves fine details such as tails, leaf patterns, and reflective surface textures, even from rear views, resulting in stable and consistent reconstructed geometry. In contrast, OpenLRM, CRM, SyncDreamer, and One-2-3-45 exhibit shape distortion such as inconsistent limb counts, texture loss, and unnatural lighting artifacts. While Wonder3D and Era3D perform better in preserving global structure, they suffer from texture drift and smoothing of local details, such as disappearing leaves or surface patterns.

Kiss3DGen also exhibits rear-view artifacts and suffers from oversmoothing of local geometric details.

These consistent differences demonstrate that UNICROSS3D not only excels in preserving global shape but also maintains fine-grained surface detail, achieving high-fidelity 3D reconstruction without geometric drift or texture distortion. This validates the effectiveness of our framework in producing detailed and structurally coherent 3D outputs across diverse and challenging scenarios.

#### 4.2.2 Text-to-3D Generation

By integrating powerful text-to-image generation models such as Stable Diffusion [RBL*22] and Imagen [SCS*22] into our framework, UNICROSS3D can be naturally extended to perform text-to-3D generation. We evaluated its effectiveness against recent state-of-the-art methods [WWC*24; JZH*24; CZY*24] for text-conditioned 3D synthesis.

We adopted CRM [WWC*24] as a baseline of the 2D-to-3D method integrated with the off-the-shelf text-to-image generator for text-to-3D generation because CRM achieved the best 3D reconstruction performance among the state-of-the-art 2D-to-3D methods in Sec. 4.2.1. For direct text-to-3D generation, we included IT3D [CZY*24] and JointDreamer [JZH*24] as comparison methods. IT3D utilizes adversarial learning in SDS-based generation to enhance visual realism and minimize geometric distortion by training the fine 3D generator and discriminator with multiview images from coarse 3D models.

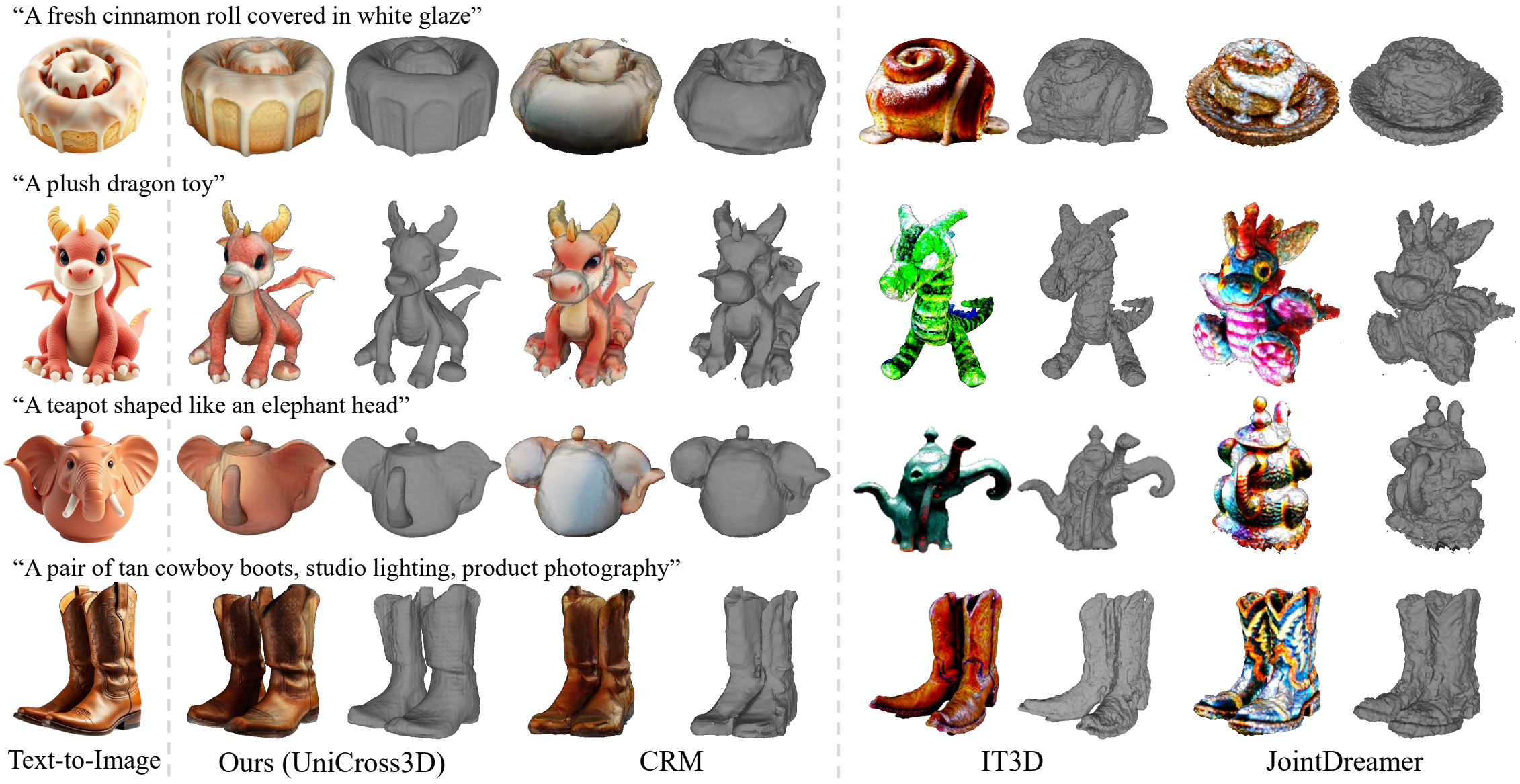

**Figure 4:** *For text-to-3D evaluation, we compare two categories of methods based on their input modality. Image-to-3D approaches such as UNICROSS3D and CRM [WWC*24] take as input the image generated by a text-to-image model (SDXL [PEL*23]) from the given text prompt (see first column). In contrast, text-to-3D methods such as IT3D [CZY*24] and JointDreamer [JZH*24] directly consume the raw text prompt as input to synthesize the 3D shape.*

**Table 2:** *Comparisons with text-to-3D methods.*

| Method | FID (↓) | CLIP score (↑) |
|---|---|---|
| JointDreamer | 99.2 | 0.349 |
| IT3D | 99.3 | 0.341 |
| CRM | 96.8 | 0.362 |
| UNICROSS3D (Ours) | **93.7** | **0.384** |

**Table 3:** *Performance comparisons with multiview images by 2D-to-3D methods.*

| Method | PSNR (↑) | SSIM (↑) | LPIPS (↓) |
|---|---|---|---|
| SyncDreamer | 18.2132 | 0.8462 | 0.1572 |
| Wonder3D | 18.0932 | 0.8995 | 0.1536 |
| Era3D | 18.4407 | 0.9088 | 0.1366 |
| UNICROSS3D (Ours) | **21.8354** | **0.9642** | **0.1061** |

JointDreamer applies Joint Score Distillation (JSD) to text prompts and multiview images, using geometry-fading and classifier-free guidance scale-switching strategies. Our UNICROSS3D framework differs from these methods in that it unifies multiview and cross-domain consistency constraints even under text-driven generation, benefiting from the explicit color-normal alignment and latent stability mechanisms described in Sec. 3. This design enables more semantically faithful and geometrically robust 3D reconstruction, even when starting from ambiguous or abstract textual prompts.

To ensure a fair comparison between image-to-3D and text-to-3D methods, we employed distinct evaluation setups for each paradigm. Specifically, for image-to-3D approaches such as UNICROSS3D and CRM, we used images generated by Stable Diffusion XL (SDXL) [PEL*23] from identical text prompts as inputs. This simulates a realistic setting where high-quality synthetic images serve as a proxy for real-world input. In contrast, for text-to-3D methods such as IT3D and JointDreamer, the text prompt was directly provided to the model to generate the 3D object without any intermediate image input. This allows us to directly evaluate the generative capabilities of text-conditioned pipelines with consistent semantic intent. For evaluation, we followed the standard protocol introduced by DreamFusion [PJBM23] and used a set of 400 diverse text prompts that span various categories and styles of objects.

Table 2 summarizes the performance comparisons in terms of FID and CLIP score. UNICROSS3D outperforms all baseline methods across both metrics, indicating its strong capability in generating text-aligned and perceptually consistent 3D content. Figure 4 presents qualitative comparisons, where UNICROSS3D produces high-fidelity 3D meshes that are structurally consistent and visually aligned with the input text, across diverse viewpoints. In the first row of Fig. 4, UNICROSS3D accurately reconstructs the shape and surface characteristics of the prompt "white glaze," capturing the specular texture and smooth geometry. In contrast, CRM, despite using the same input image, fails to reproduce the reflective surface and overall form of the object. IT3D entirely omits the "white glaze" attribute in the generated mesh, while JointDreamer only partially retains the texture but suffers from noisy geometry, resulting in a loss of structural coherence. These trends are more pronounced under complex or imaginative prompts. For instance, in the third row of Fig. 4, UNICROSS3D accurately preserves fine-grained details, such as the patterns on the elephant's ear and the boundary contour on the lid of a teapot, yielding a coherent and realistic 3D mesh. In contrast, CRM fails to capture shading information, leading to color loss and texture flattening in rear-facing regions. IT3D produces 3D meshes that are semantically misaligned with the prompt, and JointDreamer suffers from high-frequency noise in texture and shape, degrading local detail and structural integrity.

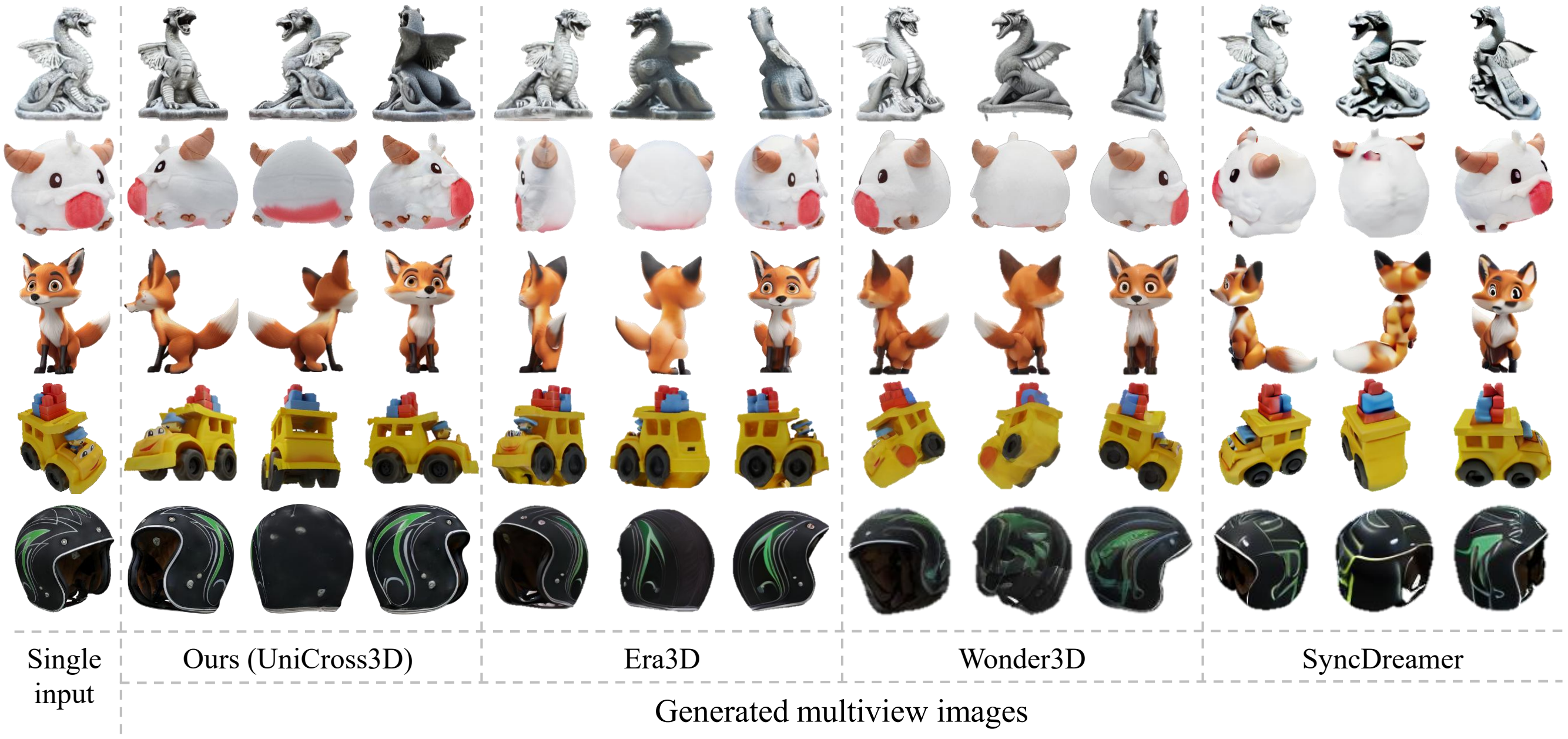

**Figure 5:** *Qualitative comparisons of multiview images generated from the single-view input image.*

These consistent results confirm that UNICROSS3D maintains robust alignment between geometry and appearance, even under diverse and challenging text inputs. By explicitly modeling semantic structure and enforcing cross-domain consistency, our method produces 3D outputs with high semantic fidelity and structural consistency, validating its effectiveness in text-to-3D generation.

#### 4.2.3 Single view-to-multiview Generation

To evaluate the quality of multiview generation, we compared our performance with recent single view-to-multiview methods [LLL*24; LGL*24; LLZ*24], introduced in Sec. 4.2.1. To ensure a fair comparison, we randomly sampled 500 objects from the GSO dataset [DFK*22] and all experimental settings for the proposed method were fixed as described in Sec. 4.1.3. We evaluated the quality of the novel view synthesis using PSNR, SSIM, and LPIPS, consistent with Sec. 4.1.2.

Table 3 summarizes the quantitative comparison results. UNICROSS3D outperforms all baseline methods across all evaluation metrics. Notably, our method achieves the highest score in SSIM, indicating that the generated multiview images exhibit strong structural consistency in both geometry and appearance across different viewpoints. This improvement can be attributed to the proposed cross-view regularization, which effectively suppresses the variance of latent representations across views. By stabilizing the latent space under varying camera poses, the model reduces geometry drift and texture inconsistencies that typically arise when synthesizing from unseen or challenging viewpoints. These results demonstrate that UNICROSS3D not only enhances overall multiview generation quality but also maintains geometric consistency and visual coherence regardless of the input viewpoint.

The visual comparisons of multiview generation results are presented in Fig. 5. The results show that UNICROSS3D produces more realistic and structurally consistent multiview images, particularly for objects with complex geometry and fine-grained texture. Specifically, in the third row of Fig. 5, UNICROSS3D successfully reconstructs detailed appearance elements such as fur texture, ear coloration, and tail shading, as well as object-specific structural features like the shape of the tail and posture. Notably, these attributes remain stable across all viewpoints, demonstrating strong multiview consistency. In contrast, Era3D fails to reconstruct the object's geometry accurately from side views, and exhibits inconsistent tail color patterns across viewpoints. Wonder3D struggles to preserve high-frequency details, such as fur texture and iris color, which appear smoothed or lost in the generated images. SyncDreamer produces unrealistic textures that fail to reflect the object's material, and suffers from geometric collapse across views, along with distorted lighting and facial structures. This trend is consistent even for objects with highly detailed textures. For instance, in the fifth row of Fig. 5, UNICROSS3D captures not only the internal surface texture but also complex patterns and reflective metallic details with remarkable consistency across viewpoints. Era3D fails to maintain these details, resulting in noticeable texture deformation depending on the view. Similarly, Wonder3D and SyncDreamer are unable to recover high-frequency patterns and surface cues, leading to view-inconsistent and structurally incoherent outputs. These results vali-date that the proposed method not only enhances consistency across generated views, but also preserves geometric alignment and visual coherence for objects with complex shapes and detailed appearances. By effectively suppressing geometry drift and texture distortion across viewpoints, UNICROSS3D achieves perceptually coherent multiview generation, where the object's structure and appearance remain consistently aligned from various viewpoints.

**Table 4:** *Quantitative ablation results for the unified cross-domain and cross-view framework.*

| Setting | PSNR (↑) | SSIM (↑) | LPIPS (↓) | CD (↓) | IoU (↑) |
|---|---|---|---|---|---|
| w/o cross-domain & cross-view | 19.2932 | 0.8971 | 0.1563 | 0.0213 | 0.4473 |
| w/o cross-domain | 20.5237 | 0.9384 | 0.1288 | 0.0156 | 0.5628 |
| w/o cross-view | 21.4628 | 0.9223 | 0.1421 | 0.0183 | 0.5347 |
| Ours (UniCross3D) | **21.8354** | **0.9642** | **0.1061** | **0.0125** | **0.6125** |

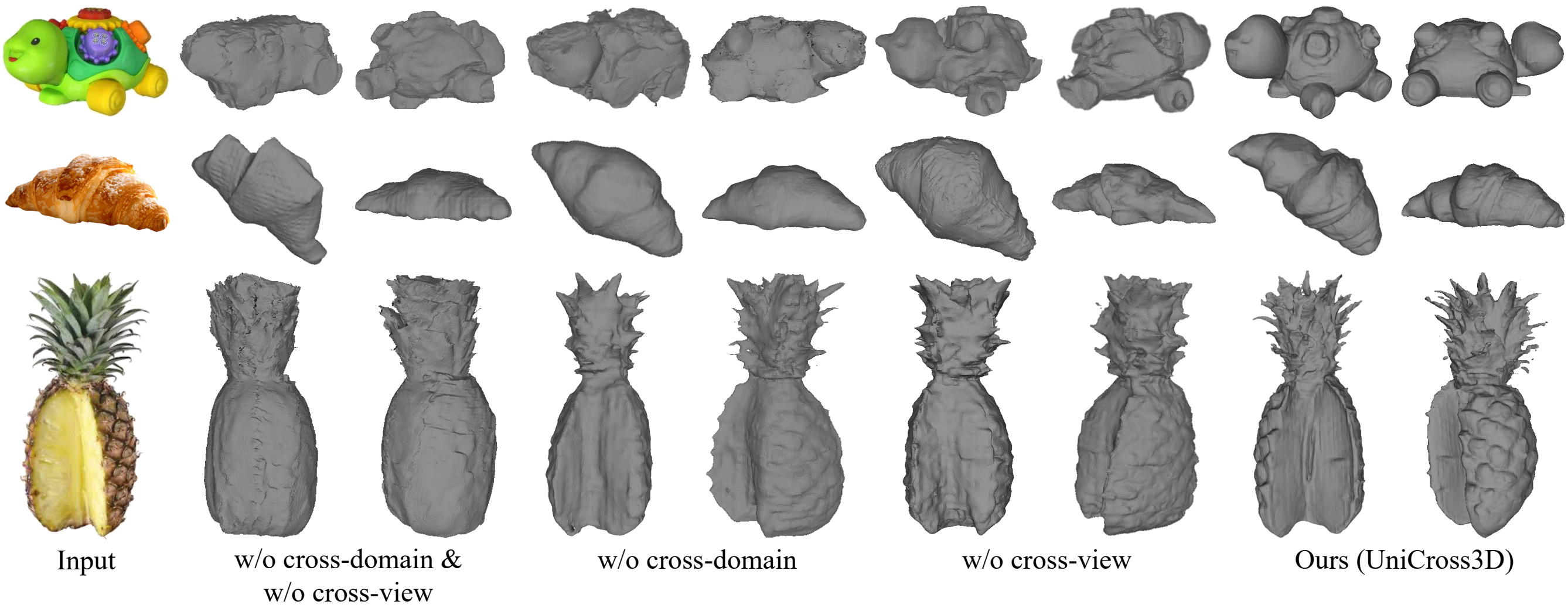

**Figure 6:** *Ablation study on unified cross-domain and cross-view framework.*

### 4.3 Ablation Studies

In this section, we conduct an ablation study to evaluate the effectiveness of the two core components of UNICROSS3D: the *cross-domain consistency mechanism* introduced in Sec. 3.4 and the *multiview latent regularization* described in Sec. 3.3. To ensure a fair comparison, all experiments were carried out in the same conditions as detailed in Sec. 4.1.3, with the only variation being the inclusion or exclusion of the respective loss terms.

We trained all models on a fixed subset of the Objaverse dataset [DSS*23] and performed evaluations on 500 randomly sampled single-view images from the GSO dataset [DFK*22] for inference. For novel view synthesis, we used Peak Signal-to-Noise Ratio (PSNR), Structural Similarity Index Measure (SSIM), and Learned Perceptual Image Patch Similarity (LPIPS). For 3D reconstruction, we used Chamfer Distance (CD) and volumetric Intersection over Union (IoU) to assess geometric accuracy. These metrics collectively measure both visual fidelity and structural correctness of the generated results. This setup allows us to isolate the contribution of each module and quantitatively validate their impact on multiview consistency and cross-domain alignment.

We evaluated the model in four different ablation settings based on the inclusion or exclusion of each consistency term. To thoroughly assess the effectiveness of the unified cross-view and cross-domain diffusion framework, we used unseen images that contain rich geometric details. The performance comparison is summarized in Table 4 and Fig. 6. The results show that applying both cross-domain consistency and multiview latent regularization yields the best performance across all quantitative metrics. It is shown that the model effectively distinguishes high-frequency textures from complex geometric structures with both components and produces stable reconstructions that preserve the object's true shape. Notably, even under challenging unseen viewpoints, the generated geometry remains undistorted and well-aligned with appearance, indicating high cross-view and cross-domain consistency. This indicates that these two components are complementary and jointly contribute to robust and coherent 3D reconstruction.

In contrast, when the cross-domain consistency term is removed, significant misalignments can be observed between appearance and geometry. Specifically, geometric structures in the input image are often misinterpreted as texture cues, leading to flattened or overly simplified normal predictions that fail to capture fine structural details—especially in regions with complex materials or high-frequency texture. These artifacts stem from a lack of mutual information between the color and normal domains during training, causing them to be learned independently. This highlights the importance of our proposed mutual information maximization objective, which enforces semantic alignment between appearance and geometry, enabling them to co-evolve during training and ultimately improving 3D reconstruction quality.

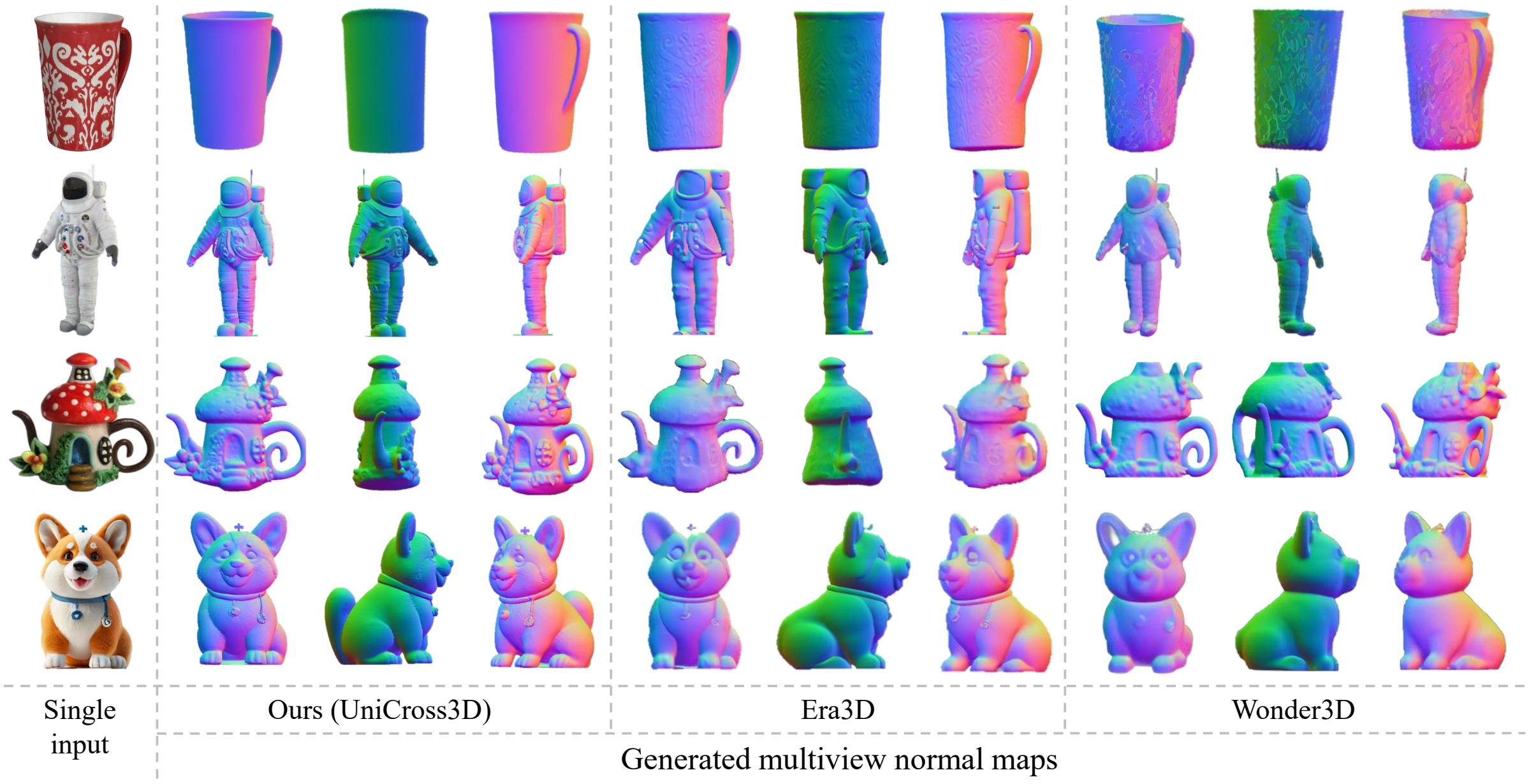

**Figure 7:** *Qualitative comparisons of multiview normal maps from the single-view input image.*

**Table 5:** *Performance comparisons of surface fidelity with 2D-to-3D methods.*

| Method | MAE (↓) |
|---|---|
| Wonder3D | 12.1428 |
| Era3D | 9.4682 |
| UNICROSS3D (Ours) | **6.2715** |

Similarly, removing the multiview latent regularization leads to a loss of shape coherence across views. In this setting, we observe geometry drift and inconsistent structure depending on the input viewpoint, indicating that the model has become overly dependent on the input viewpoint. Without regularizing the posterior variance of latent features across views, the model tends to form divergent latent codes, resulting in unstable geometry when extrapolating to unseen viewpoints. These findings validate the effectiveness of our cross-view latent regularization, which suppresses posterior variance across viewpoints during training, thereby enforcing structural consistency in the latent space. Together with our cross-domain consistency mechanism, this design enables UNICROSS3D to generate high-fidelity, geometrically coherent 3D reconstructions under diverse and ambiguous conditions.

## 4.4 Analysis of Model Capabilities

In this section, we conducted experiments to analyze the capabilities of the proposed framework by evaluating surface fidelity of reconstructed geometry in Sec. 4.4.1, and robustness of multiview generation to input viewpoint variations in Sec. 4.4.2.

### 4.4.1 Surface Fidelity Evaluation

In this section, to analyze the surface fidelity of the proposed method, we compared the performance of recent 2D-to-3D methods [LGL*24; LLL*24] that reconstruct geometry by generating multiview normal maps, including Wonder3D [LGL*24] and Era3D [LLL*24], introduced in Sec. 4.2.1.

We define surface fidelity as the model's ability to disentangle texture and geometry from the input image and accurately reconstruct the underlying surface of the object. To quantitatively assess this ability, we follow recent works [GHZ*23; WRN*24] and measure the accuracy of the generated surface normals using the Mean Angular Error (MAE). MAE calculates the average angular deviation between the predicted normals and the ground-truth normals, where lower values indicate better alignment with the true local surface orientation. For a fair comparison, we randomly sampled 500 objects from the GSO dataset [DFK*22]. All experimental settings were fixed as specified in Sec. 4.1.3.

The performance comparison for surface fidelity is summarized in Table 5. The results indicate that UNICROSS3D achieves the lowest MAE among all methods, outperforming the recent approaches. A lower MAE implies that the predicted surface normals are more closely aligned with the ground-truth normals, demonstrating that our model can effectively disentangle texture from geometry and accurately recover local surface orientation. This performance gain can be attributed to the cross-domain consistency mechanism introduced in UNICROSS3D, which promotes semantic alignment between the generated color images and normal maps. By explicitly encouraging the two domains to share consistent structural information, the model reduces errors where high-frequency textures are misinterpreted as geometry, or fine geomet-ric details are mistaken for appearance. As a result, the proposed cross-domain mutual information objective reinforces the semantic separation between appearance and shape, ultimately improving the fidelity of the generated normal maps.

**Table 6:** *Performance comparison of multiview consistency across different input viewpoints.*

| Method | PSNR (↑) | SSIM (↑) | LPIPS (↓) | CD (↓) | IoU (↑) |
|---|---|---|---|---|---|
| Era3D | 18.5132 | 0.8175 | 0.1417 | **0.0162** | **0.5066** |
| UNICROSS3D (Ours) | **20.2857** | **0.9456** | **0.1076** | **0.0131** | **0.5973** |

**Table 7:** *Sensitivity analysis of the mutual information weight $\lambda_{\mathrm{MI}}$ in the proposed method.*

| $\lambda_{MI}$ | PSNR (↑) | SSIM (↑) | LPIPS (↓) | CD (↓) | IoU (↑) |
|---|---|---|---|---|---|
| 0.01 | 20.9123 | 0.9494 | 0.1187 | 0.0149 | 0.5792 |
| 0.05 | **21.8354** | **0.9642** | **0.1061** | **0.0125** | **0.6125** |
| 0.1 | 20.6489 | 0.9438 | 0.1205 | 0.0138 | 0.5704 |

**Table 8:** *Sensitivity analysis of the latent variance weight $\lambda_{Var}$ in the proposed method.*

| $\lambda_{\mathrm{Var}}$ | PSNR (↑) | SSIM (↑) | LPIPS (↓) | CD (↓) | IoU (↑) |
|---|---|---|---|---|---|
| 0.05 | **22.0121** | **0.9670** | **0.1059** | 0.0197 | 0.5452 |
| 0.01 | 21.8354 | 0.9642 | 0.1061 | **0.0125** | **0.6125** |
| 0.005 | 20.9321 | 0.9422 | 0.1289 | 0.0146 | 0.5830 |

Figure 7 illustrates qualitative comparisons. Our method consistently recovers sharp and geometrically aligned surface normals, effectively disentangling shape and appearance. In the first row, UNICROSS3D correctly identifies complex surface patterns as part of the appearance domain and reconstructs a smooth and stable surface without introducing false geometry. In contrast, Era3D and Wonder3D misinterpret these patterns as surface shapes, resulting in bumpy and distorted geometry on what should be flat regions. This trend also appears in the opposite failure mode, where models confuse the fine geometric structure with the visual appearance. For example, in the third row of Fig. 7, UNICROSS3D successfully captures thin structural details such as flowers and window frames and faithfully encodes them in the predicted normal map. On the other hand, Era3D tends to flatten these features, leading to blurred or oversmoothed normals that obscure fine geometry. Wonder3D shows even greater degradation, with substantial geometric collapse and distortions, especially in side regions, resulting in the failure to reconstruct realistic object shapes.

These consistent results demonstrate that our method learns a more reliable mapping between appearance and geometry, preserving the integrity of both domains even under complex textures and shapes. By maintaining a clear separation between color and normal information, UNICROSS3D achieves superior surface fidelity and more accurate 3D reconstructions.

#### 4.4.2 Robustness to Input Viewpoint

In this section, we evaluated the robustness of multiview generation with respect to variations in the input viewpoint. We compared UNICROSS3D against a recent 2D-to-3D method, Era3D [LLL*24], which achieved the strongest performance in novel view synthesis in Sec. 4.2.1. To assess viewpoint robustness, we generated images of the same target viewpoint from multiple input viewpoints of the same object. The target viewpoint is defined relative to the canonical object coordinate system, allowing us to evaluate whether the model produces consistent outputs regardless of the input camera angle. Specifically, we fixed the elevation to $0^\circ$ and varied the azimuth in increments of $30^\circ$ to obtain 16 distinct input views per object. We measured consistency across these input views by computing PSNR, SSIM, and LPIPS between the images generated from the same target viewpoint.

In addition, to complement image-space consistency evaluation, we further measure 3D geometric consistency under varying input viewpoints using Chamfer Distance (CD) and IoU. For fair comparison, we randomly sampled 500 objects from the GSO dataset [DFK*22] and followed the experimental setup in Sec. 4.1.

As summarized in Table 6, UNICROSS3D consistently outperforms Era3D in all metrics. In particular, the improvements in SSIM and LPIPS indicate a higher visual and structural consistency among the generated images of the same target viewpoint.

In addition, lower CD and higher IoU in our results indicate higher geometric consistency in 3D shapes reconstructed from different input viewpoints than Era3D, beyond image-level quality alone. This suggests that our method reliably preserves the object's intrinsic geometric and appearance features, even when the input viewpoint varies significantly. This robustness can be attributed to the proposed multiview latent regularization, which minimizes the variance of latent representations across input views during training. As a result, the model learns to maintain consistent structural representations, reducing dependency on specific input viewpoints. These findings demonstrate that UNICROSS3D can generate coherent and semantically aligned multiview output, even with unseen or challenging viewpoints, without suffering from geometric distortion or appearance inconsistency.

Figure 8 illustrates qualitative comparisons of multiview image generation from different input viewpoints. Given an input image from a fixed viewpoint (column 1), each row shows generated images from UNICROSS3D (columns 2-4) and the prior method (columns 5-7) rendered at various target viewpoints (*e.g.*, $0^\circ$, $90^\circ$, $180^\circ$). These comparisons highlight whether the model produces consistent images at the same target viewpoint, regardless of input viewpoint variation. Gray boxes indicate the cases where the input and target viewpoints are identical. The results show that UNICROSS3D consistently produces semantically aligned and geometrically stable multiview images, regardless of the input viewpoint.

Specifically, in the first object of Fig. 8, UNICROSS3D preserves key geometric structures, such as the presence of wings and object thickness, across all target views with azimuthal changes, without distortion or collapse. In contrast, Era3D exhibits significant inconsistencies depending on the input view: when given a side view, essential geometric features like wings are missing or overly flattened; when given frontal or rear views, the side structure becomes distorted or unrealistic. These observations suggest that Era3D struggles to consistently capture and preserve the intrinsic shape and semantic features of the object across viewpoints. This performance gap is not limited to complex objects. Even for simpler structures, UNICROSS3D maintains a consistent and ac-curate shape across varying inputs. For example, when comparing the outputs for the same target view at azimuth 90°, UNICROSS3D produces stable and coherent poses regardless of the input angle, while Era3D shows structural deformation or tilt depending on the input view. Moreover, when a rear view is provided as input, UNICROSS3D is still able to reconstruct partial semantic details such as the face of the object, while Era3D fails to recover any meaningful features.

These results highlight the effectiveness of our view-consistent objective, which enforces latent representation alignment across viewpoints. By minimizing view-dependent variation in the latent space, UNICROSS3D learns to model object structure and semantics uniformly across inputs, avoiding overfitting to specific views and producing consistent outputs. In summary, UNICROSS3D demonstrates strong robustness against unseen or challenging viewpoints and excels at preserving both geometric and semantic fidelity in multiview generation, for both simple and complex objects alike.

## 4.5 Hyperparameters Analysis

In this section, we analyze the sensitivity of the proposed method with respect to the mutual information parameter $\lambda_{\text{MI}}$ and the multiview latent variance parameter $\lambda_{\text{Var}}$ in (18).

### 4.5.1 Sensitivity Analysis of $\lambda_{\text{MI}}$.

We evaluated the sensitivity of the proposed method to the cross domain consistency by varying the weight parameter $\lambda_{\text{MI}}$ in (18). We trained our model on a subset of the Objaverse dataset [DSS*23] for the 2D-to-3D generation task and evaluated its performance under different values of $\lambda_{\text{MI}}$ to assess how strongly the mutual information objective influences the model's ability to align appearance and geometry. To ensure a fair comparison, the other hyperparameters and experimental setups were fixed to Sec. 4.1.3. We randomly sampled 500 single-view images from the GSO dataset [DFK*22], and evaluated the 2D-to-3D generation quality using PSNR, SSIM, and LPIPS for novel view synthesis, and CD and IoU for 3D reconstruction. We conducted experiments using three values of $\lambda_{\text{MI}}$: 0.01, 0.05, and 0.1. The performance comparison for the 2D-to-3D generation task is summarized in Table 7.

The results show that setting $\lambda_{\text{MI}} = 0.05$ yields the best empirical performance across both novel view synthesis and 3D reconstruction metrics. This indicates that the mutual information objective introduced in UNICROSS3D effectively enhances semantic alignment between the color and normal domains, thereby overcoming key limitations of prior methods, such as misinterpreting high-frequency texture patterns in the input image as geometric structures, or conversely, treating actual surface geometry as mere texture variation.

The trained model with too small $\lambda_{\text{MI}}$ (*i.e.*, 0.01) insufficiently learns the semantic alignment between color and normal domains. As a result, it struggles to distinguish high-frequency texture from true geometric variation. This is especially evident in regions with sharp curvature or complex structure, where surface normals tend to become overly flattened. In such cases, even when color images are visually accurate, the corresponding normal maps may become distorted or underrepresented, leading to inaccurate geometric reconstruction. While this misalignment may not severely degrade novel view synthesis, it ultimately limits the structural fidelity of the textured 3D mesh. In contrast, when $\lambda_{\text{MI}}$ is excessively large (*i.e.*, 0.1), the model places disproportionate emphasis on aligning the color and normal domains. Although this improves per-view semantic consistency, it can degrade multiview coherence. We observe cases where shading in one view becomes inconsistent with actual geometry when observed from other views, leading to reconstruction artifacts such as unstable surface contours or inconsistent fine details across viewpoints. This effect negatively impacts both novel view synthesis and the overall geometric integrity of the reconstructed shape.

These results suggest that enforcing sufficient cross-domain semantic alignment enables the model to better distinguish between visual textures in the input image and the underlying geometric structure. They also underscore the importance of properly weighting the mutual information objective to ensure both accurate appearance-geometry alignment and coherent multiview reconstruction. Notably, this trend is consistently observed in our experiments on text-to-3D generation tasks as well. In all settings, we found that setting $\lambda_{\text{MI}} = 0.05$ yields the most stable and robust performance across a wide range of prompts and object categories.

### 4.5.2 Sensitivity Analysis of $\lambda_{\text{Var}}$.

We evaluated the sensitivity of the proposed method with respect to the multiview latent variance weight $\lambda_{\text{Var}}$ in (18), using a fixed subset of the Objaverse dataset [DSS*23] across different parameter values. All other hyperparameters were fixed to the values specified in Sec. 4.1.3 to ensure a fair comparison. We randomly sampled 500 single-view images from the GSO dataset [DFK*22], and evaluated the quality of 2D-to-3D generation using PSNR, SSIM, and LPIPS for novel view synthesis, and CD and IoU for 3D reconstruction. We conducted experiments with three values of $\lambda_{\text{Var}}$: 0.005, 0.01, and 0.05. The comparison results are summarized in Table 8.

The results show that setting $\lambda_{\text{Var}} = 0.01$ achieves the best performance across quantitative metrics for both novel view synthesis and 3D reconstruction. This demonstrates that the multiview latent regularization introduced in UNICROSS3D effectively suppresses the variance of latent representations across different viewpoints, thereby overcoming key limitations of prior methods such as geometry distortions and appearance inconsistencies caused by variations in input viewpoint or noise seed.

Consequently, when $\lambda_{\text{Var}}$ is set too low (*i.e.*, 0.005), the regularization across latent representations from different views becomes insufficient. As a result, the latent features become highly sensitive to variations in the input view, such as camera parameters, viewpoint, or noise seed. This leads to inconsistent latent encodings for the same object under different conditions and increases the posterior variance across views. Particularly under unseen viewpoints, the misaligned latent representations deviate from the training distribution, resulting in geometric drift, shading artifacts, or lighting inconsistencies. Such instability degrades the generalization performance of the model throughout the entire pipeline, affecting both view synthesis and 3D reconstruction. In contrast, setting $\lambda_{\text{Var}}$ too high (*i.e.*, 0.05) overly constrains the latent space, forcing different views to collapse into nearly identical representations. While this may improve view consistency metrics such as PSNR, SSIM, and LPIPS, as shown in Table 8, the model fails to capture fine-grained details and viewpoint-specific variations in lighting and texture. As a result, it tends to produce repetitive and overly flattened outputs across different views. This reduced diversity leads to the loss of local structure and visual richness and ultimately decreases the quality of 3D reconstruction, resulting in a significant increase in CD and a decrease in IoU, which indicate that the reconstructed shapes deviate from the ground truth geometry and exhibit reduced volumetric fidelity. The resulting meshes often exhibit oversimplified geometry or repeated patterns, limiting the expressiveness and realism of the textured output.

These results demonstrate that an appropriate level of multiview latent regularization is essential to balance generalization and representational diversity. In particular, setting $\lambda_{\text{Var}} = 0.01$ achieves this balance effectively, stabilizing latent features across views while preserving sufficient variation for high-fidelity reconstruction. We also observed consistent trends in other tasks, such as text-to-3D generation, where $\lambda_{\text{Var}} = 0.01$ similarly produced competitive and stable results across diverse prompts and object categories.

## 5 Conclusion

We presented UniCross3D, a unified cross-view and cross-domain diffusion framework for high-fidelity single-image 3D generation. Our method addresses two primary challenges: viewpoint inconsistency and appearance-geometry misalignment. To this end, we proposed (1) multiview latent regularization to prevent drift across views, and (2) cross-domain mutual information maximization to align color and normal predictions. Extensive experiments on 2D-to-3D and text-to-3D tasks showed that UniCross3D achieves strong performance in view synthesis, 3D reconstruction accuracy, and surface normal quality. The model effectively reconstructed complex geometry and fine textures, even from challenging inputs. A limitation of our approach is its reliance on an external NeRF-based backend for final mesh reconstruction, which may affect the quality of the final 3D output. Nonetheless, our results showed that enforcing consistency and alignment in the generative process yields significantly improved 3D reconstruction, even without explicit 3D supervision. We believe UniCross3D offers a promising step toward physically grounded and semantically consistent 3D generation.

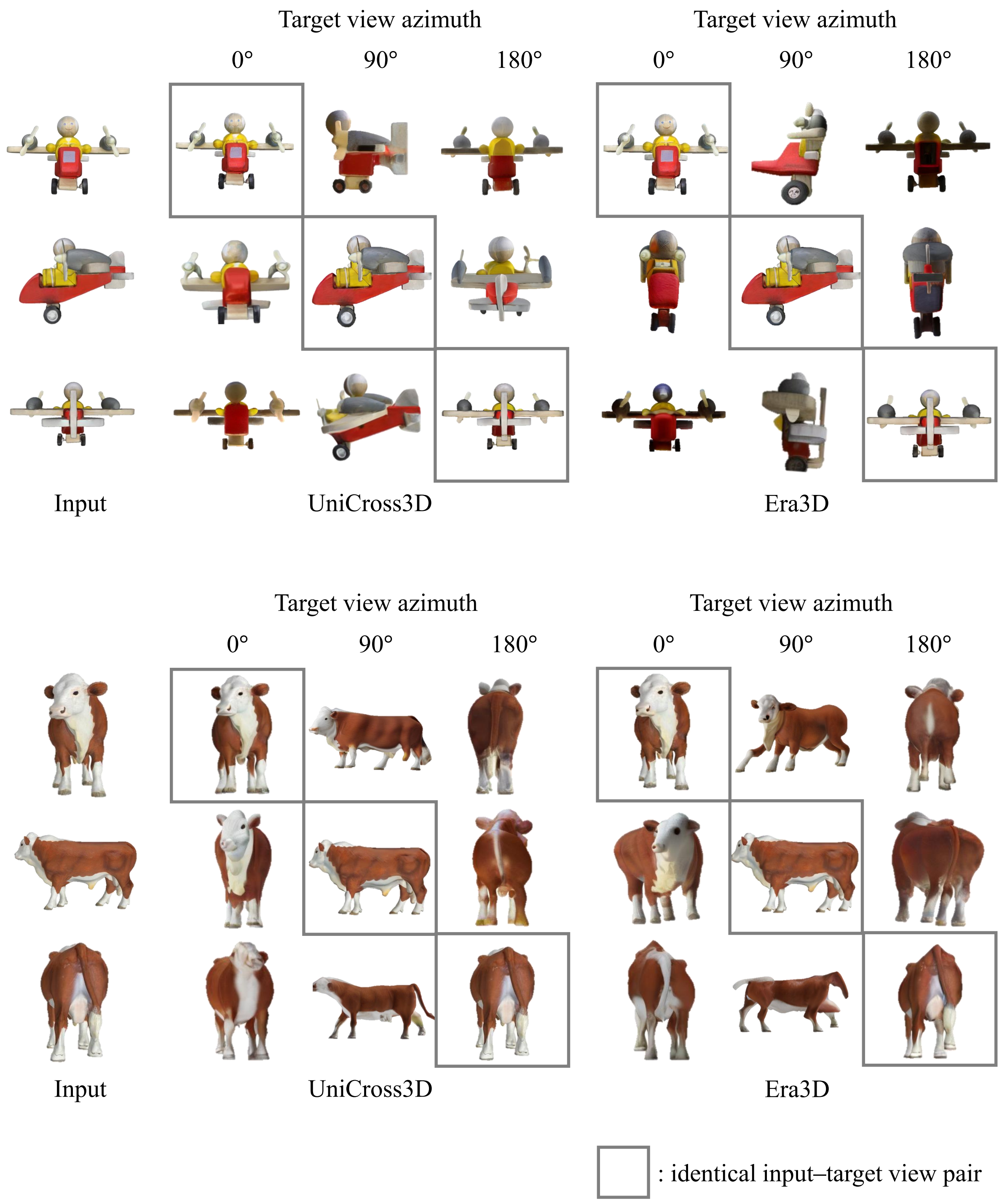

**Figure 8:** *Qualitative comparisons of multiview images generated from different input viewpoints. Each row shows the generated multiview images for a fixed target viewpoint, generated from different input views. The first column shows the input image, followed by the results from* UNICROSS3D *(columns 2-4) and Era3D (columns 5-7). Gray boxes indicate cases where the input and target viewpoints are identical.*